\documentclass{llncs}
\usepackage{llncsdoc}
\usepackage{graphicx}
\usepackage{subfigure}
\begin{document}
\title{Adaptive Forgetting Factor Fictitious Play}
\author{Michalis Smyrnakis \thanks{Michalis Smyrnakis is supported by The Engineering and Physical Sciences Research Council EPSRC (grant number EP/I005765/1).}\inst{1}  \and David S. Leslie\inst{2}}
\institute{School of Physics and Astronomy University of Manchester,UK. \email{michalis.smyrnakis@manchester.ac.uk} \and Department of Mathematics University of Bristol, UK. \\ \email{david.leslie@bris.ac.uk}}

\maketitle

\begin{abstract}
It is now well known that decentralised optimisation can be formulated as a potential game, and game-theoretical learning algorithms can be used
to find an optimum. One of the most common learning techniques in game theory
is fictitious play. However fictitious play is founded on an implicit assumption that
opponents' strategies are stationary. We present a novel variation of fictitious play
that allows the use of a more realistic model of opponent strategy. It uses a heuristic approach, from the online streaming data literature, to adaptively update the
weights assigned to recently observed actions. We compare the results of the proposed algorithm with those of stochastic and geometric fictitious play in a simple
strategic form game, a vehicle target assignment game and a disaster management
problem. In all the tests the rate of convergence of the proposed algorithm was
similar or better than the variations of fictitious play we compared it with. The
new algorithm therefore improves the performance of game-theoretical learning in
decentralised optimisation.
\end{abstract}

\section{Introduction}
\label{intro}
Decentralised optimisation is a crucial component of sensor networks \cite{cn,lesser}, disaster management \cite{dm}, traffic control \cite{tf} and scheduling  \cite{scp}. In each of these
domains a combination of computational and communication complexity render centralised optimisation approaches intractable. It is now well known that many
decentralised optimisation problems can be formulated as a potential game \cite{wlu,autonomous,lesliearchie}. Hence the optimisation problem can be recast in terms of finding a Nash
equilibrium of a potential game. An iterative decentralised optimisation algorithm
can therefore be considered a type of learning in games algorithm, and vice versa.

Fictitious play is the canonical example of learning in games \cite{learning_in_games}. Under fictitious play each player maintains some beliefs about his opponents' strategies, and
based on these beliefs he chooses the action that maximises his expected reward.
The players then update their beliefs about opponents' strategies after observing
their actions. Fictitious play converges to Nash equilibrium for certain kinds of
games \cite{learning_in_games,fp4} but in practice this convergence can be very slow. This is because
it implicitly assumes that the other players use a fixed strategy in the whole game
by giving the same weight to every observed action.

In \cite{cj} this problem was addressed by using particle filters to predict opponents'
strategies. The drawback of this approach is the computational cost of the particle
filters that render difficult the application of this method in real life applications.

In this paper we propose an alternative method which uses a heuristic rule
to adapt the weights of opponents' strategies by taking into account their recent
actions. We observe empirically that this approach reduces the number of steps
that fictitious play needs to converge to a solution, and hence the communications
overhead between the distributed optimisers that is required to find a solution to
the distributed optimisation problem. In addition the computational demand of
the proposed algorithm is similar to the classic fictitious play algorithm.

The remainder of this paper is organised as follows. We start with a brief description of game theory, fictitious play and stochastic fictitious play. Section \ref{afffp}
introduces adaptive forgetting factor fictitious play (AFFFP). The impact of the algorithm's parameters on its performance is studied in Section \ref{test}. Section \ref{results} presents
the results of AFFFP for a climbing hill game, a vehicle target assignment game
and a disaster management simulation scenario. We finish with a conclusion.

\section{Background}
\label{background}
In this section we introduce the relationship between potential games and decentralised optimisation, as well as the classical fictitious play learning algorithm.

\subsection{Potential games and decentralised optimisation}
A class of games which maps naturally to the decentralised optimisation framework
is strategic form games. The elements of a strategic form game are \cite{games1}

\begin{itemize}
\item a set of players ${1,2,\ldots,I}$,
\item a set of actions $s^{i}\in S^{i}$ for each player $i \in I$,
\item a set of joint actions, $s=(s^{1}, s^{2}, \ldots\ , s^{I}) \in S^{1} \times S^{2}\times \ldots\times S^{I}=S $,
\item a payoff function $u^{i}:S \rightarrow \mathbf{R}$ for each player $i$, where $u^{i}(s)$ is the utility that player $i$ will gain after a
specific joint action $s$ has been played.
\end{itemize}
We will often write
$s=(s^{i},s^{-i})$, where $s^{i}$ is the action of Player $i$ and
$s^{-i}$ is the joint action of Player $i$'s opponents.

The rules that the players use to select the action that they will play in a game are called strategies. A player $i$ chooses his actions according to a pure
strategy when he selects his actions by using a deterministic rule. In the cases that he chooses an action based on a
probability distribution then he acts according to a mixed
strategy. If we denote the set of all the
probability distributions over the action space $S^{i}$ as $\Delta^{i}$, then a
mixed strategy of player $i$  is an element 
$\sigma^{i} \in \Delta^{i}$. We define $\Delta$ as the set product of all
$\Delta^{i}$, $\Delta=\Delta^{1} \times \ldots \times
\Delta^{I}$. Then the  joint mixed strategy $\sigma=(\sigma^{1}, \ldots , \sigma^{I})$, is defined as an element
of $\Delta$ and we will often write $\sigma=(\sigma^{i},
\sigma^{-i})$ analogously to $s=(s^{i},s^{-i})$. We will denote
the expected utility a player $i$ will gain if he chooses a
strategy $\sigma^{i}$ (resp.\ $s^{i}$), when his opponents choose
the joint strategy $\sigma^{-i}$ as
$u^{i}(\sigma^{i},\sigma^{-i})$ (resp.\
$u^{i}(s^{i},\sigma^{-i})$).

Many decision rules can be used by the players to choose their actions in
a game. One of them is to choose their actions from a set of mixed strategies
that maximises their expected utility given their beliefs about their opponents'
strategies. When Player $i$'s opponents' strategies are $\sigma^{-i}$ then the best response
of player $i$ is defined as:
\begin{equation}
BR^{i}(\sigma^{-i})= \mathop{\rm argmax}_{\sigma^{i}\in \Delta^{i}} \quad
u^{i}(\sigma^{i},\sigma^{-i}).
\label{eq:bestres}
\end{equation}

Nash \cite{nash}, based on Kakutani's fixed point theorem, showed that
every game has at least one equilibrium. This equilibrium is a
strategy $\hat{\sigma}$ that is a fixed point of the best response
correspondence, $\hat{\sigma}^{i} \in BR^{i}(\hat{\sigma}^{-i}) \forall i$. Thus when a
joint mixed strategy $\hat{\sigma}$ is a Nash equilibrium then
\begin{equation}
u^{i}(\hat{\sigma}^{i},\hat{\sigma}^{-i})\geq u^{i}(s^{i},\hat{\sigma}^{-i}) \qquad \textrm{for all } i, \textrm{ for all }  s^{i} \in S^{i}.
\label{eq:nashutil}
\end{equation}
 
Equation (\ref{eq:nashutil}) implies that if a strategy
$\hat{\sigma}$ is a Nash equilibrium then it is not possible for a player to
increase his utility by unilaterally changing his strategy.
When all the players in a game select equilibrium actions using pure
strategies then the equilibrium is referred as pure strategy
Nash equilibrium.

A particularly useful category of games for multi-agent decision problems is
the class of potential games \cite{fp4,lesliearchie,autonomous}. The utility function of an exact potential
game satisfies the following property:

\begin{equation}
u^{i}(s^{i},s^{-i})-u^{i}(\tilde{s^{i}},s^{-i})=\phi(s^{i},s^{-i})-\phi(\tilde{s^{i}},s^{-i})
\label{pot_games}
\end{equation}
where $\phi$ is a potential function and the above equality stands
for every player $i$, for every action $s^{-i}\in S^{-i}$, and for
every pair of actions $s^{i}$, $\tilde{s^{i}} \in S^{i}$. The potential function depicts the changes in the players' payoffs when they unilaterally change
their actions. Every potential game has at least one pure strategy Nash equilibrium
\cite{fp4}. There may be more than one, but at any equilibrium no player can increase
their reward, therefore the potential function, through a unilateral deviation.

Wonderful life utility \cite{wlu,autonomous} is a method to design the individual utility functions of a potential game such that the global utility function of a decentralised
optimisation problem acts as the potential function. Player $i$'s utility when a joint
action $s = (s^i , s^{-i} )$ is performed, is the difference in global utility obtained by the
player selecting action $s^i$ in comparison with the global utility that would have
been obtained if $i$ had selected an (arbitrarily chosen) reference action $ s^i_0$:

\begin{equation}
u^{i}(s^{i},s^{-i})=u_{g}(s^{i},s^{-i})-u_{g}(s^{i}_{0},s^{-i})
\label{eq:wlu}
\end{equation}
where $u_g$ is the global utility function. Hence the decentralised optimisation problem can be cast as a potential game, and any algorithm that is proved to converge
to Nash equilibria will converge to a joint action from which no player can increase
the global reward through unilateral deviation.

\subsection{Fictitious Play}
\label{fictitious_play}
Fictitious play is a widely used learning technique in game theory. In fictitious play each player chooses
his action according to the best response to his beliefs about opponent's
strategy.

Initially each player has some prior beliefs about the strategies
that his opponents use to choose actions. The
players, after each iteration, update their beliefs about their opponents' strategy and
play again the best response according to their beliefs. More
formally in the beginning of a game players maintain some arbitrary non-negative initial weight functions
$\kappa_{0}^{j}$, $j=1, \ldots, I$ that are updated using the formula:
\begin{equation}
\kappa_{t}^{j}(s^{j}) = \kappa_{t-1}^{j}(s^{j})+I_{s^j_t=s^j}
\label{eq:kappa}
\end{equation}
for each $j$, where
$I_{s^j_t=s^j}=\left\{\begin{array}{cl}1&\mbox{if
$s^j_t=s^j$}\\0&\mbox{otherwise.}\end{array}\right.$. \\The mixed
strategy of opponent $j$ is estimated from the following formula:

\begin{equation}
\sigma_{t}^{j}(s^{j})=\frac{\kappa_{t}^{j}(s^{j})}{\sum_{s^j}{\kappa_{t}^{j}(s^{j})}}.
\label{eq:fp1p}
\end{equation}
Equations (\ref{eq:kappa}) and (\ref{eq:fp1p}) are equivalent to: 

\begin{equation}
\sigma_{t}^{j}(s^{j})=\left(1-\frac{1}{t^j}\right)\sigma^j_{t-1}(s^j) + \frac{1}{t^j}I_{s^j_t=s^j}
\label{eq:fp}
\end{equation}
where $t^j=t+\sum_{s^j\in S^j}\kappa^j_0(s^j)$. Player $i$ chooses an action which maximises his expected payoffs given his beliefs about his opponents' strategies.

The main purpose of a learning algorithm like fictitious play is to converge
to a set of strategies that are a Nash equilibrium. For classic fictitious play (\ref{eq:fp})
it has been proved \cite{learning_in_games} that if $\sigma$ is a strict Nash equilibrium and it is played at
time $t$ then it will be played for all further iterations of the game. Also any steady
state of fictitious play is a Nash equilibrium. Furthermore, it has been proved
that fictitious play converges for 2 $\times$ 2 games with generic payoffs \cite{fp2}, zero sum
games \cite{fp1}, games that can be solved using iterative dominance \cite{fp3} and potential
games \cite{fp4}. There are also games where fictitious play does not converge to a
Nash equilibrium. Instead it can become trapped in a limit cycle whose period is
increasing through time. An example of such a game is Shapley's game \cite{shapley}.

A player $i$ that uses the classic fictitious play algorithm uses best responses to his beliefs to choose his actions, so he chooses his actions $s^{i}$ from his
pure strategy space $S^{i}$. Randomisation is allowed only in the case that
players are indifferent between his available actions, but it is
very rare in generic payoff strategic form games for a player to
be indifferent between the available actions \cite{fad}.
Stochastic fictitious play is a variation of fictitious play
where players use mixed strategies in order to choose actions. This variation was originally introduced to
allow convergence of players' strategies to a mixed strategy Nash
equilibrium \cite{learning_in_games} but has the additional advantage
of introducing exploration into the process.

The most common form of smooth best response of Player $i$, $\overline{BR}^{i}(\sigma^{-i})$, is  the following \cite{learning_in_games}:
\begin{equation}
\overline{BR}(\sigma^{-i})(s^{i})=\frac{\exp(u^{i}(s^{i},\sigma^{-i})/\xi)}{\sum_{\tilde{s}^{i}}\exp(u^{i}(\tilde{s}^{i},\sigma^{-i})/\xi)}
\label{eq:smbr}
\end{equation}
where $\xi$ is the randomisation parameter. When the value of
$\xi$ is close to zero, a $\overline{BR}$ is close to $BR$ and players exploit their action space, whereas
large values of $\xi$ result in complete randomisation
\cite{learning_in_games}.

Stochastic fictitious play is a modification of fictitious play under which Player $i$ uses $\overline{BR}^{i}(\sigma^{-i}_{t})$ to randomly select an action instead of selecting a best response action $BR^{i}(\sigma^{-i}_{t})$. We reinforce the fact that the difference  between classic 
fictitious play and stochastic fictitious play is in the decision
rule the players use to choose the actions. The updating rule (\ref{eq:fp}) that is used to update the beliefs of the opponents' strategies are the same in both algorithms.

When Player $i$ uses equation (\ref{eq:fp}) to update the beliefs about opponents' strategies he treats the environment of the game as stationary and implicitly assumes
that the actions of the players are sampled from a fixed probability distribution \cite{learning_in_games}. Therefore recent observations have the same weight as initial ones. This approach leads to poor adaptation when other players change their strategies.

A variation of fictitious play that treats the opponents' strategies as dynamic
and places greater weights on recent observations while we calculate each action's
probability is geometric fictitious play, introduced in \cite{learning_in_games}. According to this variation of fictitious play the estimation of each opponent's probability to play an
action $s^{j}$ is evaluated using the formula:
\begin{equation}
\sigma_{t}^j(s^{j})=(1-z)\sigma_{t-1}^{j}(s^{j})+ zI_{s^j_t=s^j}
\label{gfp}
\end{equation}
where $z \in (0, 1) $ is a constant.

In Section \ref{afffp} we introduce a new variant of fictitious play in which the constant $z$ is automatically adapted in response to the observations of opponent strategy.

\section{Adaptive forgetting factor fictitious play}
\label{afffp}

The objective of players when they maintain beliefs $\sigma^{-i}_{t}$ is to estimate the mixed
strategy of opponents. However consider streaming data where in each time step
a new observation arrives and it belongs to one of $J$ available classes \cite{Anag}. When
the objective is to estimate the probability of each class given the observed data,
this objective can be expressed as the fitting of a multinomial distribution to the
observed data stream. If a fixed multinomial distribution over time is assumed,
then its parameters can be estimated using the empirical frequencies of the previously observed data. This is exactly the strategy estimation described in Section \ref{fictitious_play}. But in real life applications it is rare to observe
a data stream from a constant distribution. Hence there is a need for the learning algorithm to adapt to the distribution that the data currently follow. This is
similar to iterative games, where we expect that all players update their strategies
simultaneously.

An approach that is widely used in the streaming data literature to handle
changes in data distributions is forgetting. This suggests that recent observations
have greater impact on the estimation of the algorithm's parameters than the older
ones. Two methods of forgetting are commonly used: window based methods and
resetting. Salgado et.al \cite{Salgado} showed that when abrupt changes (jumps) are detected
then the optimal policy is to reset the parameters of the algorithm. In the case of
smooth changes (drift) in the data stream's distribution, a solution is to use only
a segment of the data (a window). The simplest form of window based methods
uses a specific segment size constantly; there are also approaches that adaptively
change the size of the window but they are more complicated. Some examples of
algorithms that use window based methods are \cite{Black2,MCS,Aggar}.

Another method is to introduce forgetting, which is also used in geometric
fictitious play (\ref{gfp}), to discount the old information by giving higher weights to the
recent observations. When the discount parameter is fixed it is necessary to know
a priori the distribution of data and the way that they evolve through time due
to the fact that we must choose the forgetting factor in advance. In addition the
performance of the approximation when there are changes that result from a jump
or non-constant drift is poor for a fixed forgetting factor. For those reasons this
methodology has serious limitations.

A more sophisticated solution is the use of a forgetting factor that takes into
account the recent data and the previously estimated parameters of the model and
adapts to observed changes in the data distribution. Such a forgetting factor was
proposed by Haykin \cite{Haykin} in the case of recursive least squares filtering problems.
In \cite{Haykin} the objective was the minimisation of the mean square error of a cost
function that depends on an exponential weighting factor $\lambda$. This forgetting factor
is then recursively updated using gradient descent of the forgetting factor, $\lambda$,
with respect to the residual errors of the algorithm. Anagnostopoulos \cite{Anag} proposed
a generalisation of this method in the context of online streaming data from a
generalised linear model according to which the forgetting factors are adaptively
changed by using gradient ascent of the log-likelihood of the new data point.

In the streaming data context, after $t$ time intervals we observe a sequence of data $x_{1},\ldots,x_{t}$ and we fit a model $f(\theta_{t}|x_{1:t})$, where $\theta_{t}$ are the model's parameters at time $t$. Note that the parameters of the model, $\theta_{t}$, depend on the observed data stream $x_{1:t}$ and the forgetting factors $\lambda_{t}$. Since the estimated model parameters depend on $\lambda_{t}$ we will write $\theta_t(\lambda_{t})$. The log-likelihood of the data that will arrive at time $t+1$, $x_{t+1}$, given the parameters of the model at time $t$ will be denoted as $\mathcal{L}(x_{t+1};\theta_{t}(\lambda_{t}))$. Then the update of the forgetting factor $\lambda_{t+1}$ can be expressed as:

\begin{equation}
\lambda_{t+1}=\lambda_{t}+\gamma \frac{\partial \mathcal{L}(x_{t+1};\theta_{t}(\lambda_{t}))}{\partial \lambda}
\label{eq:affan}
\end{equation}
where $\gamma$ is the learning rate parameter of the gradient ascent algorithm.

As in \cite{Anag}, we can apply the forgetting factor of equation (\ref{eq:affan}) in the case of
fitting a multinomial distribution to streaming data. This will result a new update
rule that players can use, instead of the classic fictitious play update rule (\ref{eq:fp}), to
maintain beliefs about opponents' strategies.

In classic fictitious play the weight function (\ref{eq:kappa}) places the same weight on every observed action. In particular $\kappa_{t}^{j}(s^{j})$ denotes the number of times that player $j$ has played the action $s^{j}$ in the game. To introduce forgetting the impact of the previously observed actions in the weight function will be discounted by a factor $\lambda_{t-1}$. Such a weight function can be written as:
\begin{equation}
\kappa_{t}^{j}(s^{j})=\lambda_{t-1}^{j} \kappa_{t-1}^{j}(s^{j})+ I_{s^{j}_{t-1}=s^{j}}
\label{eq:afffp_kappa}
\end{equation}
where $I_{s^{j}_{t-1}=s^{j}}$ is the same identity function as in (\ref{eq:fp}). To normalise we set $n_{t}=\sum_{s^{j}\in S^{j}}\kappa_{t}^{j}(s^{j})$. From the definition of $\kappa_{t}^{j}(s^{j})$ we can use the following recursion to evaluate $n^{j}_{t}$
\begin{equation}
n_{t}^{j}=\lambda_{t-1}^{j}n_{t-1}^{j}+1. 
\label{eq:affp_ni}
\end{equation} 
Then player $i$'s beliefs about his opponent $j$'s probability of playing action $s^{j}$ will be:
\begin{equation}
\sigma_{t}^{j}(s^{j})=\frac{\kappa_{t}^{j}(s^{j})}{n_{t}^{j}}.
\label{eq:affp_sigma}
\end{equation}

Similarly to the case of geometric fictitious play $0<\lambda_{t} \leq1$. Moreover when the value of $\lambda_{t}$ is close to zero this results in very fast adaptation and when $\lambda_{t}=0$ the players are myopic, and thus they respond to the last action of their opponents. On the other hand when $\lambda_{t}=1$ this results in the classic fictitious play update rule. 

From this point on in we will only consider inference over a single opponent mixed
strategy in fictitious play. In the case of multiple opponents separate estimates are formed
identically and independently for each opponent. We will therefore drop all dependence on player $i$, and
write $s_{t}$, $\sigma_{t}$ and $\kappa_{t}(s)$ for the opponent's action,
strategy and weight function respectively.

The value of $\lambda$ should be updated in order to have adaptive forgetting factors. Initially we have to evaluate the log-likelihood of the recently observed action $s_{t}$ given the beliefs of the opponents strategies. The log-likelihood is of the following form:
\begin{equation}
\mathcal{L}(s_{t}; \sigma_{t-1})=\ln \sigma_{t-1}(s_{t})
\label{eq:affp_likli}
\end{equation}
When we  replace $\sigma_{t-1}(s_{t})$ with its equivalent from (\ref{eq:affp_sigma}) the log-likelihood can be written as:
\begin{equation}
\mathcal{L}(s_{t}; \sigma_{t-1})=\ln \big( \frac{\kappa_{t-1}(s_t)}{n_{t-1}} \big)= \ln \kappa_{t-1}(s_t) - \ln n_{t-1}
\label{eq:affp_loglikli}
\end{equation}

In order to estimate the update of $\lambda$, equation (\ref{eq:affan}), the evaluation of the log-likelihood's derivative with respect to $\lambda$ is required. The terms $\kappa_{t}$ and $n_{t}$ both depend on $\lambda$. Hence the derivative of (\ref{eq:affp_loglikli}) is:

\begin{equation}
\frac{\partial\mathcal{L}(s_{t}; \sigma_{t-1})}{\partial \lambda}= \frac{1}{\kappa_{t-1}(s_{t})}\frac{\partial}{\partial \lambda} \kappa_{t-1}(s_{t}) - \frac{1}{n_{t-1}}\frac{\partial}{\partial \lambda} n_{t-1}
\label{eq:affp_likli_der}
\end{equation}
Note that $\kappa_{t}(s_{t})=\lambda_{t-1} \kappa_{t-1}(s_{t})+ I_{s^{j}_{t-1}=s^{j}}$,  so

\begin{equation}
\frac{\partial}{\partial \lambda} \kappa_{t}(s) \vert_{\lambda=\lambda_{t-1}}= \kappa_{t-1}(s)+ \lambda_{t-1}\frac{\partial}{\partial \lambda} \kappa_{t-1}(s) \vert_{\lambda=\lambda_{t-1}}	
\label{eq:kappa_tonos}
\end{equation}
and similarly 

\begin{equation}
\frac{\partial}{\partial \lambda} n_{t}\vert_{\lambda=\lambda_{t-1}}=n _{t-1}+ \lambda_{t-1}\frac{\partial}{\partial \lambda} n_{t-1}\vert_{\lambda=\lambda_{t-1}}	
\label{eq:ni_tonos}
\end{equation}
We can use equations (\ref{eq:kappa_tonos}) and (\ref{eq:ni_tonos}) to recursively estimate $\frac{\partial}{\partial \lambda} \kappa_{t-1}(s)$ for each $s$ and $\frac{\partial}{\partial \lambda} n_{t-1}$ and hence calculate $\frac{\partial}{\partial \lambda} \mathcal{L}(s_{t}; \sigma_{t-1})$. Summarising we can evaluate the adaptive forgetting factor $\lambda_{t}$ as follows:
\begin{equation}
\lambda_{t}=\lambda_{t-1}+ \gamma \Big( \frac{1}{\kappa_{t-1}(s)}\frac{\partial}{\partial \lambda} \kappa_{t-1}(s) - \frac{1}{n_{t-1}}\frac{\partial}{\partial \lambda} n_{t-1} \Big)
\label{eq:affp_update}
\end{equation}

To ensure that $\lambda_{t}$ remains in (0, 1) we truncate it to this interval whenever it leaves.

After updating their beliefs players can choose their actions by choosing either a best response to their beliefs of their opponents strategies or a smooth best response. Table \ref{tab:affpact} summarises the algorithm of adaptive forgetting factor fictitious play.

\begin{table}
\begin{center}
\begin{tabular}{|p{12cm}|}
\hline
At time $t$, each player carries out the following
\begin{enumerate}
\item Updates the weights $\kappa_{t}^{j}(s^{j})=\lambda_{t-1}^{j} \kappa_{t-1}^{j}(s^{j})+ I_{s^{j}_{t-1}=s^{j}}$

\item Update $\frac{\partial}{\partial \lambda} \kappa_{t-1}^{j}(s)$ and $\frac{\partial}{\partial \lambda} n_{t-1}^{j}$  using equations (\ref{eq:kappa_tonos}) and (\ref{eq:ni_tonos})

\item Based on the weights of step 1 each player updates his beliefs about his opponents strategies using  $\sigma_{t}^{j}(s^{j})=\frac{\kappa_{t}^{j}(s^{j})}{n_{t}^{j}}$, where $n_{t}^{j}=\lambda_{t-1}^{j}n_{t-1}^{j}+1$.

\item Choose an action based on the beliefs of step 3 according either to best response, $BR$, or to smooth best response $\overline{BR}$

\item Observe opponent's action $s_{t}^{j}$

\item Update the forgetting factor using: $\lambda_{t}^{j}=\lambda_{t-1}^{j}+ \gamma \Big(\frac{1}{\kappa_{t-1}^{j}(s)}\frac{\partial}{\partial \lambda} \kappa_{t-1}^{j}(s) - \frac{1}{n_{t-1}^{j}}\frac{\partial}{\partial \lambda} n_{t-1}^{j}\Big)$

\end{enumerate}
\\
\hline
\end{tabular}
\caption{Adaptive forgetting factor fictitious play algorithm}
\label{tab:affpact}
\end{center}
\end{table}

\section{AFFFP parameters}
\label{test}

The adaptive rule that we choose to update the forgetting factor $\lambda$ is
based on the gradient ascent algorithm. It is well known that different initial
values of an algorithm's parameter and learning rates $\gamma$ can lead to poor
results of the gradient ascent algorithm \cite{cbishop}. This is because very
small values of $\gamma$  lead to poor exploration of the space and thus the
gradient ascent algorithm can be trapped in an area where the solution is not
optimal, whereas large values of $\gamma$ can result in big jumps that will lead
the algorithm away from the area of the optimum solution. Thus we should
evaluate the performance of adaptive forgetting factor fictitious play for
different combinations of the step size parameter $\gamma$ and initial values
$\lambda_{0}$. 

We employed a toy example, where a single opponent chooses his actions using a
mixed strategy which has a sinusoidal form, a situation which corresponds to
smooth changes in the data distribution of online streaming data. The opponent
uses a strategy of the following form over the $t=1, 2, \ldots 1000$ iterations
of the game: $\sigma_{t}(1)=\frac{cos\frac{{2\pi t}}{\beta}+1}{2}=1-\sigma_t(2)$,
where $\beta=1000$. We repeated this example 	100 times for each combination
of $\gamma$ and $\lambda_{0}$. Each time we measured the mean square error of
the estimated strategy against the real one. The range of $\gamma$ and
$\lambda_{0}$ was $10^{-6}\leq \gamma \leq 10^{-1} $ and $10^{-1} \leq
\lambda_{0} \leq 1$ respectively.

\begin{figure}
\centering
 \subfigure[Contour plot of mean square error when the range of $\gamma$ and
$\lambda$ is \mbox{$10^{-6}\leq \gamma \leq 10^{-1}$}  and \mbox{$10^{-1} \leq
\lambda \leq 1$}
respectively.]{\label{fig:msea}\includegraphics[scale=0.20]{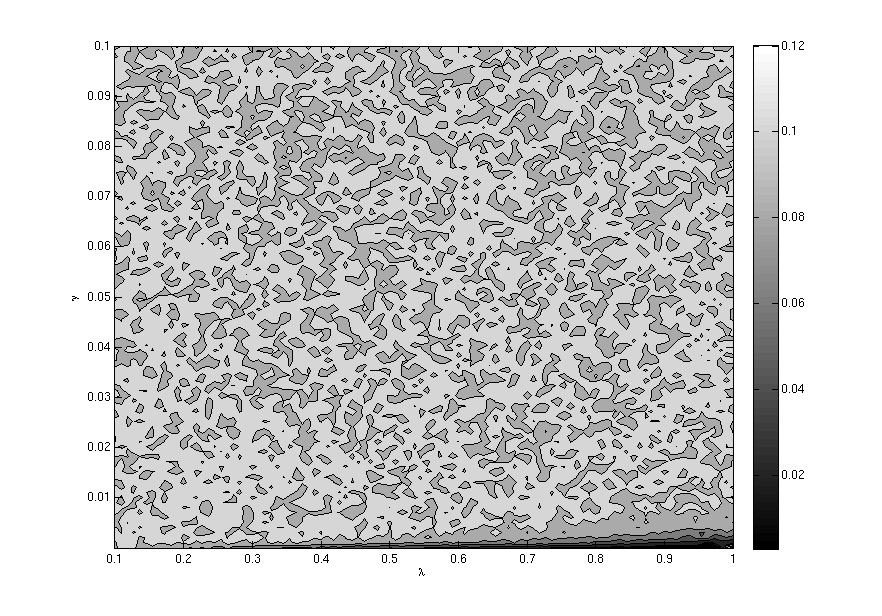}}
\subfigure[Contour plot of mean square error when the range of $\gamma$ and
$\lambda$ is \mbox{$10^{-6}\leq \gamma \leq 5 \times 10^{-3}$}  and
\mbox{$0.6\leq \lambda \leq 1$} respectively]{\label{fig:mseb}\includegraphics[scale=0.20]{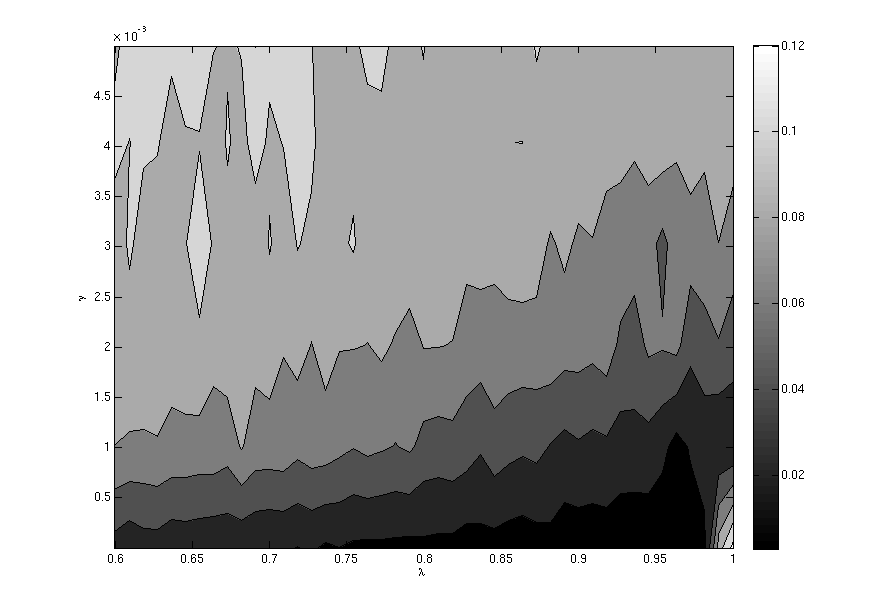} }
  \caption{Contour plot of mean square error}
  \label{fig:mse}
\end{figure}

The average mean square error for all the combinations of $\gamma$ and
$\lambda_{0}$ is depicted on Figure \ref{fig:mse}. The mean square error is
minimised in the dark area of the contour plot. In Figure \ref{fig:msea} we
observe that when $\gamma$ is less than $10^{-3}$ and $\lambda_{0}$ is greater
than $0.6$ the mean square error is minimised. In Figure \ref{fig:mseb} we
reduce the range of $\gamma$ and $\lambda_{0}$ to be  $10^{-6}\leq \gamma \leq 5
\times 10^{-3} $ and $0.6 \leq \lambda_{0} \leq 1$, respectively. Values of
$\lambda_{0}$ greater than 0.75 result in estimators with small mean square
error, for certain values of $\gamma$,  and as the value of $\lambda_0$ approaches
0.95 so it is minimised for a wider range of learning rates. We also observe
that when $\lambda_{0}$ is greater than 0.98 and as we approach 1 then the mean
square error increases. This suggests that for values of $\lambda_{0}$ greater
than $0.98$ the value of $\lambda$ appoaches $1$ very fast and thus the
algorithm behaves like the classic fictitious play update rule. In contrast when
$\lambda_{0}$ is less than 0.75 then we introduce big discounts to the
previously observed actions from the beginning of the game and the players
easily use strategies that are reactions to their opponent's randomisation.  In
addition, independently from the initial value of $\lambda_0$, when the learning
rate $\gamma$, is greater than 0.001 the algorithm results in poor estimations
of the opponent's strategy. This is because for $\gamma$ greater than 0.001 the
step that  a player moves towards the maximum of the log-likelihood is very
large and that results in values of $\lambda$ which are close either to zero or
to one. So the player uses either the classic fictitious play update rule  or
responds to his opponent's last observed action. 

\begin{figure}
\centering
\hspace{-45pt}
\includegraphics[scale=0.25]{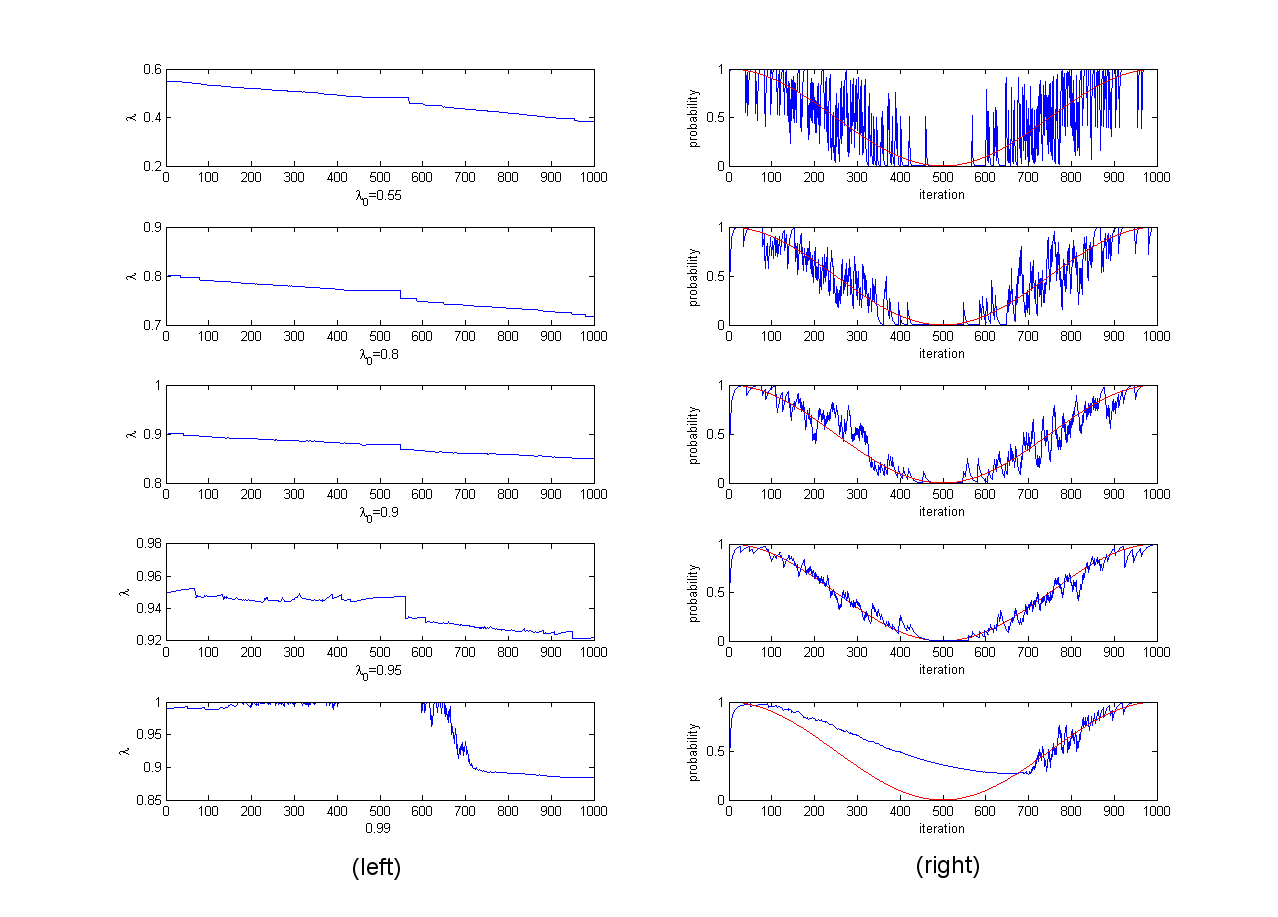}
\caption{ Evolution of $\lambda$ and tracking of $\sigma_{t}(1)$ when the true
strategies are mixed when $\gamma$ is fixed. The pre-specified strategy of
the opponent and its prediction are depicted as the red and blue
line respectively.}
\label{fig:affpexample1}
\end{figure}

\begin{figure}
\centering
\hspace{-45pt}
\includegraphics[scale=0.25]{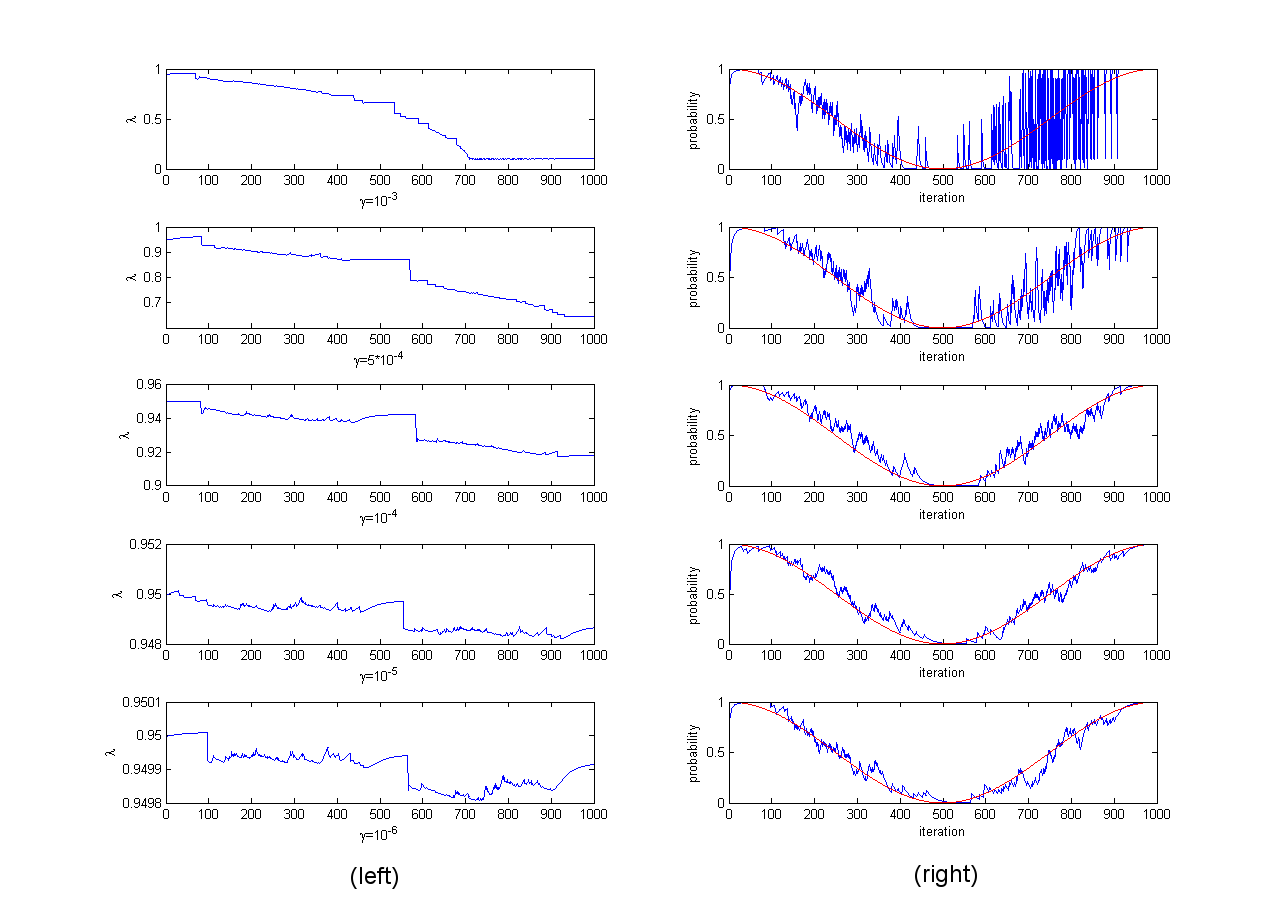}
\caption{ Evolution of $\lambda$ and tracking of $\sigma_{t}(1)$ when the true
strategies are
mixed when $\lambda_{0}$ is fixed. The pre-specified strategy of
the opponent and its prediction are depicted as the red and blue
line respectively}
\label{fig:affpexample2}
\end{figure}

We further examine the relationship between the  performance of adaptive
forgetting factor fictitious play and the sequence of $\lambda$'s in the drift
toy example with respect to the initial values of the parameters $\lambda_{0}$
and $\gamma$ . We use two instances of the drift toy example. In the first one
we set a fixed value of parameter $\gamma=10^{-4}$ and examine the performance
of our algorithm and the evolution of $\lambda$ during the game for different
values  of $\lambda_{0}=\{0.55,0.8,0.9,0.95,0.99\}$. In the second one we fix
$\lambda_{0}$ and examined the results of the algorithm for different values of
$\gamma=\{10^{-6},10^{-5},5\cdot10^{-4},10^{-4},10^{-3}\}$. Figures
\ref{fig:affpexample1} and \ref{fig:affpexample2} depict the results of the case
of fixed $\gamma$  and $\lambda_{0}$, respectively. Each row of these figures
consists of two plots for the same set of parameters, $\gamma$ and
$\lambda_{0}$. The left figure shows the evolution of $\lambda$ during the game
and the right one depicts the pre-specified strategy of the opponent and its
corresponding prediction.

As we observe in Figure \ref{fig:affpexample1}, when we set $\lambda_{0}=0.55$
the tracking of the opponent's strategy was affected by his randomisation. The
value of $\lambda$ constantly decreases which results in giving higher weights
to the recently observed actions even if they are a consequence of
randomisation. When we increase the value of $\lambda_{0}$ to 0.8 the results
are improving. When we increase $\lambda_{0}$ to 0.90 or 0.95 the resulting sequence
of $\lambda$'s does not affect the tracking of opponent's strategy. On the other
hand when we increase the value of $\lambda_{0}$ to 0.99 the value of $\lambda$
is very close to 1 for many iterations which result in poor approximation for
the same reasons that the classic fictitious play update rule  fails to capture
smooth changes in opponent's strategy.  When $\lambda_{0}$ is decreased to 0.9 the
approximation of opponent's strategy improves significantly.

Figure \ref{fig:affpexample2} depicts the results when $\lambda_{0}=0.95$ for
different values of parameter $\gamma$. We observe that high values of $\gamma$
($\gamma=10^{-3}, 5\cdot 10^{-4}$) result in big changes in the value of
$\lambda$ and that affects the quality of the approximation. On the other hand
when we use very small values of $\gamma$, $\gamma=10^{-5}$, or $\gamma=
10^{-6}$, it leads to very small deviations from $\lambda_{0}$. The good 
approximation results of opponent's strategy that we observe for those two
values of $\gamma$ are because of the initial value $\lambda_{0}$. In this
scenario if we fix the value of $\lambda=0.95$ during the whole game we will
also have a good approximation. But in real life applications it is impossible
to choose so efficiently the value of $\lambda_{0}$. When $\gamma=10^{-4}$ we
observe changes in the values of $\lambda$, which are not so sudden as the ones
for $\gamma=10^{-3}$ or $\gamma=5\cdot10^{-4}$, that lead to good approximation
of opponent's strategy. 

\begin{figure}
\centering
\hspace{-45pt}
\includegraphics[scale=0.25]{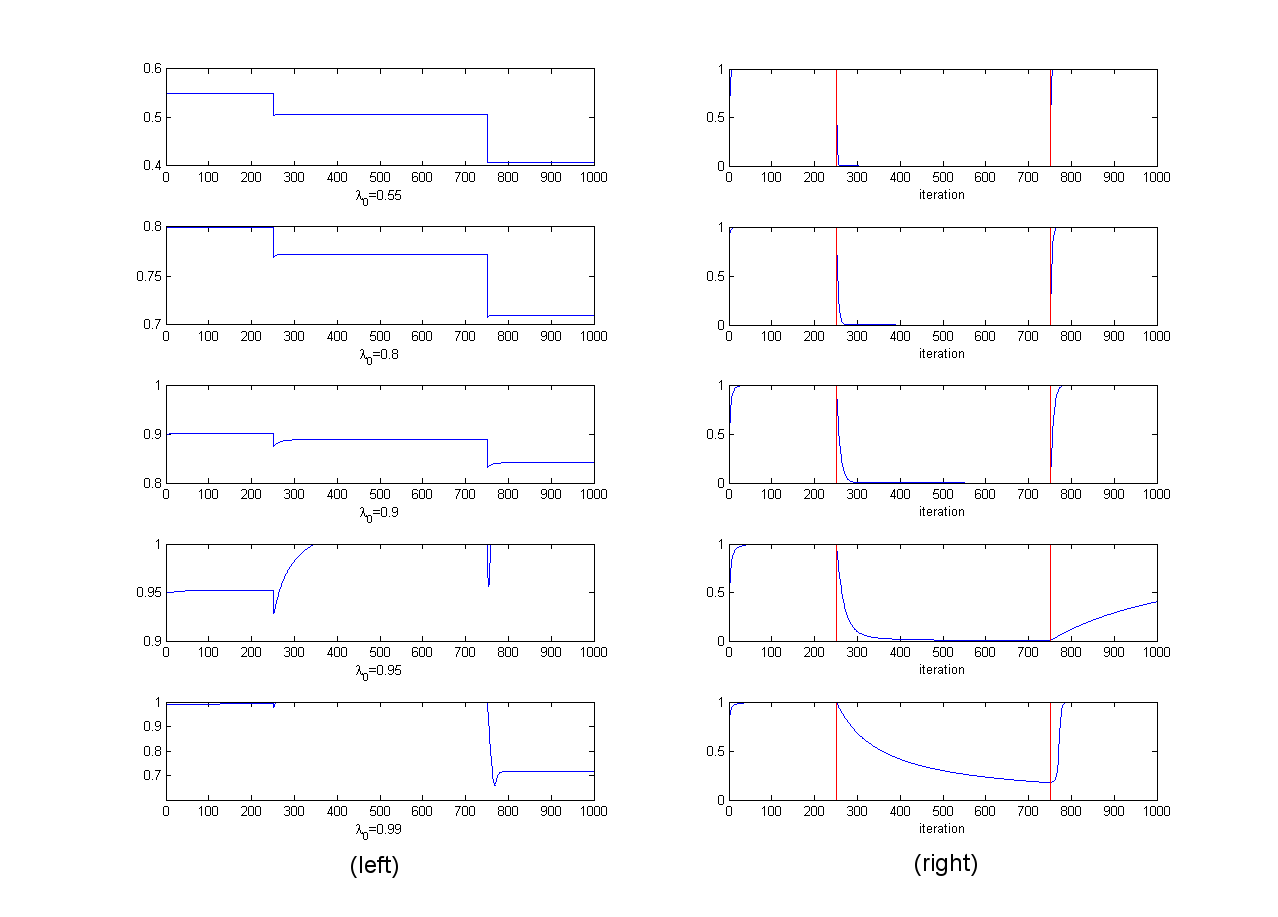}
\caption{ Evolution of $\lambda$ and tracking of $\sigma_{t}(1)$ when the true
strategies are
pure when $\gamma$ is fixed. The pre-specified strategy of
the opponent and its prediction are depicted as the red and blue
line respectively.}
\label{fig:affpexample3}
\end{figure}

\begin{figure}
\centering
\hspace{-45pt}
\includegraphics[scale=0.25]{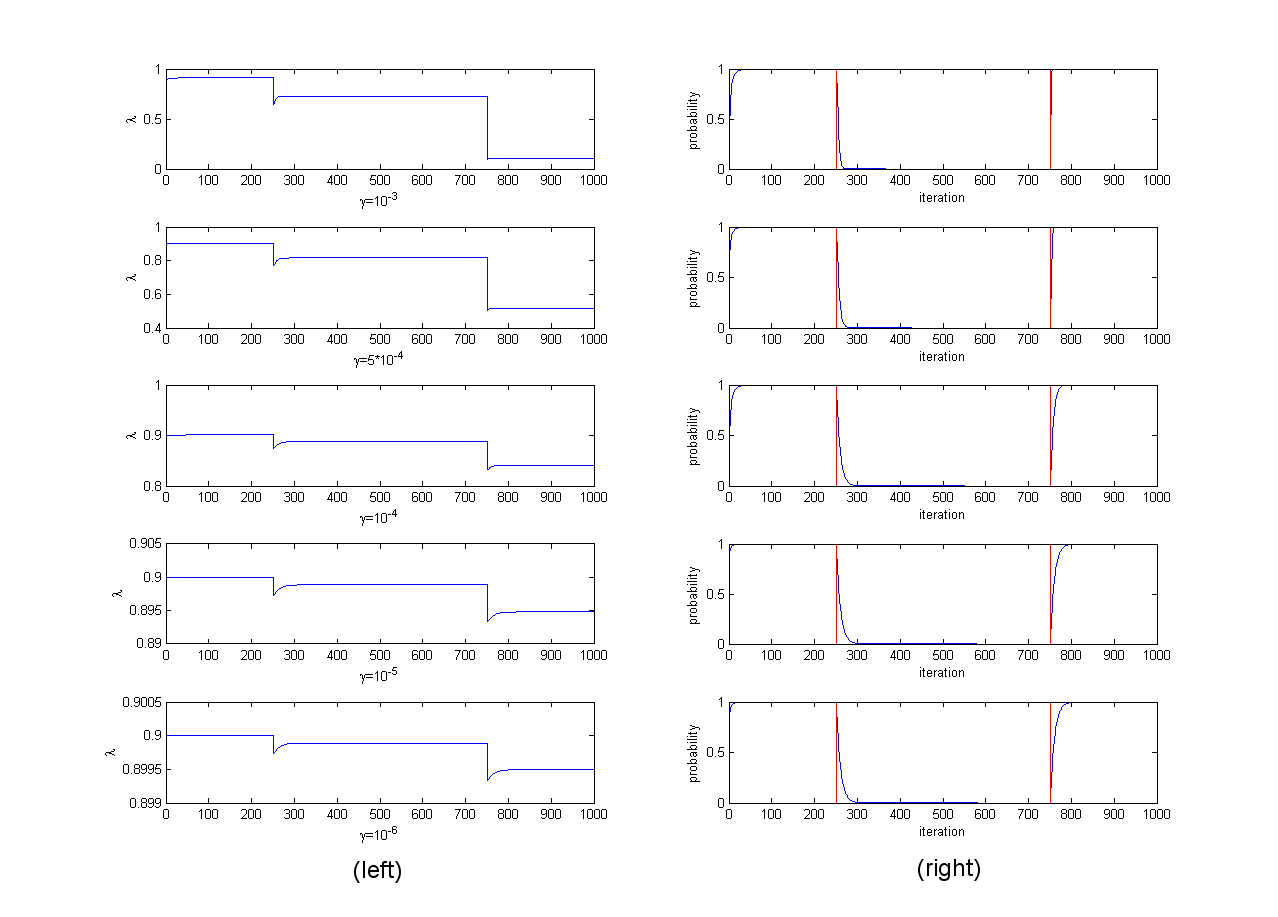}
\caption{ Evolution of $\lambda$ and tracking of $\sigma_{t}(1)$ when the true
strategies are
pure when $\lambda_{0}$ is fixed. The pre-specified strategy of
the opponent and its prediction are depicted as the red and blue
line respectively.}
\label{fig:affpexample4}
\end{figure}

We also performed simulations for a case where jumps occur. In this example we
used a game of $1000$ iterations and two available actions, 1 and 2. The
opponent played action 1  with probability $\sigma_{t}^{2}(1)=1$ during the
first 250 and the last 250 iterations of the game and for the remaining
iterations of the game $\sigma_{t}^{2}(1)=0$. The probability of the second
action can be calculated by  using $\sigma_{t}^{2}(2)=1 - \sigma_{t}^{2}(1)$. 
The results for the case of fixed $\gamma$  and $\lambda_{0}$ are depicted in
Figures \ref{fig:affpexample3} and \ref{fig:affpexample4}, respectively.

When abrupt changes occur the different values of $\gamma$ do not affect the
performance of the algorithm as we observe in Figure \ref{fig:affpexample4}. On
the contrary the initial value of $\lambda$ affects the estimation of the jumps
in the opponent's strategies. As we observe in Figure \ref{fig:affpexample3} the
opponent's strategy approximation and the evolution of $\lambda$ are similar
when $\lambda_{0}$ is equal to 0.55, 0.8 and 0.9 respectively. In those three
cases when a jump is observed, there is a drop in the value of $\lambda$, then
$\lambda$ slightly increases and finally it remains constant until the next jump
occurs. 

The sequences of $\lambda$ and the approximation results of the last two cases,
$\lambda_{0}=0.95$ and $\lambda_{0}=0.99$ are different from the 3 cases we
described above. In both of them the opponent's strategy tracking is good at the
first 250 iterations, but afterwards these two examples have the opposite
behaviour. In the example where $\lambda_{0}=0.95$ the opponent's strategy is
correctly approximated when $\sigma_{t}(1)=0$. Because of the high weights of
the previously observed actions, the likelihood needs a large number of
iterations to become constant and thus $\lambda$ becomes equal to 1. Then the
adaptive forgetting factor fictitious play process becomes identical to classic
fictitious play and fails to adapt to the second jump. When $\lambda_{0}$ is
equal to 0.99 adaptive forgetting factor fictitious play fails to adapt the
estimation of opponent's strategy to the first jump. But when the second jump
occurs, the likelihood of action 1 is small since action 2 is played for 500
consecutive iterations, and a drop in the value of $\lambda$ is observed which
resulted in adaptation to the change of opponent's strategy.

By taking into account the above results we observe that $\gamma=10^{-4}$
and $0.8 \leq \lambda_{0}\leq 0.9$ leads to useful approximations when we
consider cases where both smooth and abrupt changes in opponent's strategy are
possible to happen. In the remainder of the article we set $\gamma=10^{-4}$ and $\lambda_{0}=0.8$.

%%%%%%%%%%Results

\section{Results}
\label{results}
\subsection{Climbing hill game}
We initially compared the performance of the proposed algorithm with the
results of geometric and classic stochastic fictitious play in a three player
climbing hill game. This game which is depicted in Table \ref{table:chl2}, generalises the climbing hill game that was presented in \cite{clh} and exhibits a long best response path from the risk-dominant joint mixed action (D,D,U) to the Nash equilibrium.

\begin{table}
%\flushleft
\begin{center}
%\hspace*{-0.5cm}
\begin{tabular}{c|c|c|c}
\begin{tabular}{c} ~\\ \footnotesize U\\ \footnotesize M\\ \footnotesize D\\ ~
\end{tabular}
&
\begin{tabular}{ccc}
\small U &\small M &\small D \\    \hline
 \small   0&\small 0&\small 0\\
  \small  0&\small 50&\small 40\\
 \small   0&\small 0&\small 30\\      \hline
 \small   &U&
 \end{tabular}
&
 \begin{tabular}{ccc}
\small U &\small M &\small D \\ \hline
\small -300&\small 70&\small 80\\
\small -300&\small 60&\small 0\\
\small 0&\small 0&\small 0\\     \hline \small &M&
\end{tabular}
 &
 \begin{tabular}{ccc}
 \small U &\small M & \small D \\    \hline
\small $\mathbf{100}$& \small -300&\small 90\\
\small 0& \small 0& \small 0\\
\small 0&\small 0&\small 0\\           \hline
 \small &D&
 \end{tabular}

\end{tabular}

%\begin{tabular}{|c|c|c|c||c|c|c||c|c|c|}
%\hline
% &U&M&D&U&M&D&U&M&D\\ \hline
% U&0&0&0&-300&70&80&100&-300&90\\ \hline
% M&0&50&40&-300&60&0&0&0&0\\ \hline
% D&0&0&30&0&0&0&0&0&0\\ \hline
% \end{tabular}
% \begin{tabular}{c c c c c c c c c c}
%U&&&&&M&&&&D\\
%\end{tabular}
 
\caption{ Climbing hill game with three players.  Player 1 selects rows, Player
2 selects columns, and Player 3 selects the matrix. The global reward depicted in the matrices, is recived by all players. The unique Nash equilibrium is in bold.}
\label{table:chl2}
\end{center}
\end{table}

We present the results of 1000 replications of a learning episode
of 1000 iterations for each game. For each
replication of 1000 iterations we computed the mean payoff. After
the end of the 1000 replications the overall mean of the 1000
payoff means was computed. 

The value of the learning parameter $z$ of geometric fictitious
play  was set to $0.1$. We selected this value of $z$ on the pemise that the algorithm has the best results in the tracking experiment with pre-specified opponents strategy that we used in Section \ref{test} to select the parameters of AFFFP. Thus we will use this learning rate ia all the simulations that we present in the rest of this article. For all algorithms we used smooth best responses (\ref{eq:smbr}) with randomisation parameter $\xi$ in the smooth best response 
function equal to 1, allowing the same randomisation for all algorithms.

\begin{figure}
\centering
\includegraphics[scale=0.3]{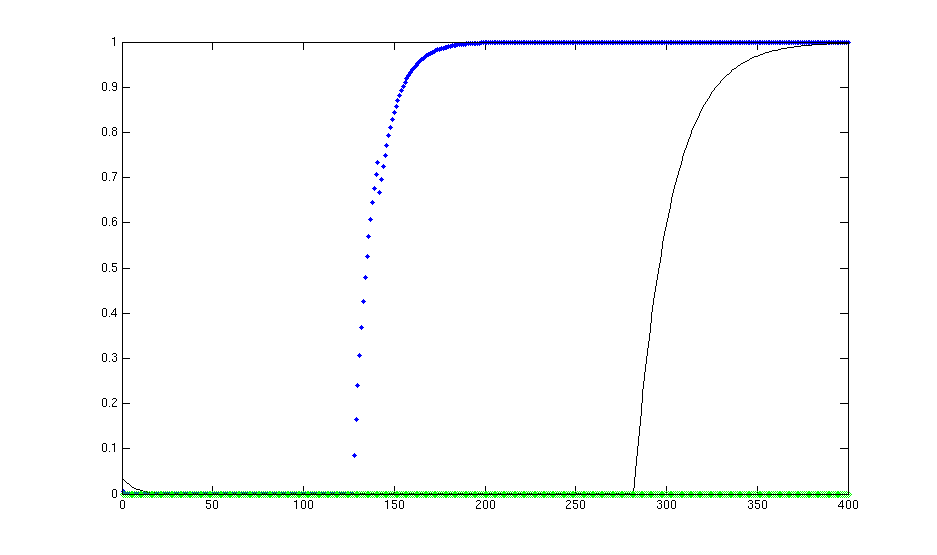}
\caption{Probability of playing the (U,U,D) equilibrium for one run of each of
AFFFP (blue dot line), geometric  fictitious
play
(black solid line) and stochastic fictitious play (green diamond line) for the
three player climbing hill game.}
\label{fig:clhil2}
\end{figure}

Adaptive forgetting factor fictitious play performed better than both geometric
and
stochastic fictitious play. The overall mean global payoff was 95.26 for AFFFP  whereas the respective payoffs for geometric
and stochastic fictitious play were 91.7 and 70.3. Stochastic
fictitious play didn't converge to the Nash equilibrium after 1000
replications. Also when we are concerned about the speed of
convergence the proposed variations of fictitious play outperform geometric
fictitious
play. This can be seen if we reduce the iterations of the game to
200. Then the overall mean payoffs of AFFFP is 90.12  when for geometric
fictitious play it is 63.12. This is because adaptive forgetting factor
fictitious play requires approximately  100 iterations to reach the Nash
equilibrium, when geometric fictitious play needs at least 300. This difference
is depicted in Figure \ref{fig:clhil2}.

\subsection{Vehicle target assignment game}
We also compared the performance of the proposed variation of fictitious play
against the results of geometric fictitious play in the vehicle target
assignment game that is described in \cite{autonomous}. In this game agents 
should coordinate to achieve a common goal which is to maximise the total
value of the targets that are destroyed. In particular in a
specific area we place $I$ vehicles and $J$ targets. For each vehicle $i$,
its available actions are simply the targets that are available to
engage. Each vehicle can choose only one target to engage but a
target can be engaged by many vehicles. The probability that
player $i$ has to destroy a target $j$ is $p_{ij}$ if it chooses
to engage target $j$. We assume that the probability each agent has to
destroy a target is independent of the actions of the other agents, and the target is destroyed if any one agent succesfully destroyes it, so the
probability a target $j$ is destroyed by the vehicles that
engage it is $1- \prod_{i: s^{i}=j}{(1-p_{ij})}$. Each of the
targets has a different value $V_j$. The expected utility that is
produced from the target $j$ is the product of its value $V_j$ and
the probability it has to be destroyed  by the vehicles that
engage it. More formally we can express the utility that is
produced from target $j$ as:
\begin{equation}
U_{j}(s)=V_{j}(1- \prod_{i:s^{i}=j}{(1-p_{ij}}))
\label{eq:target_utility}
\end{equation}
The global utility is then the sum of the utilities of each
target:
\begin{equation}
u_{g}(s)= \sum_{j}{U_{j}(s)} \label{eq:global_utility}.
\end{equation}

Wonderfull life utility was used to to evaluate each vehicle's payoff. Then the utility that a vehicle $i$ receives after engange a target $j$, $s^{i}=j$, is 
\begin{equation}
 u_{i}(s^{i}, s^{-i})=U_{j}(s^{i}, s^{-i})-U_{j}(s^{i}_{0}, s^{-i})=
\end{equation}
where $s_{0}^{i}$ was set to be the greedy action of player $i$: $s_{0}^{i}=\displaystyle \mathop{\rm argmax}_{j} \textrm{ } V_{j}p_{ij}.$ 

In our simulations we used thirty
vehicles and thirty targets that were placed uniformly at random in a unit
square. The probability of a vehicle $i$ to destroy a target $j$
is proportional to the inverse of its distance from this target
$1/d_{ij}$. The values of the targets are independently sampled from a uniform
distribution with range in [0 100].

The vehicles had to ``negotiate'' with
the other vehicles (players) for a fixed number of negotiation
steps before they choose a target to engage. A negotiation step begins with each
player choosing a target to
engage and it ends by the agents exchanging this information with
the others, and updating their beliefs about their opponents'
strategies based on this information. The target that each vehicle will
choose in the game will be his action at the final negotiation step.

\begin{figure}%
\begin{center}
	\includegraphics[scale=0.3]{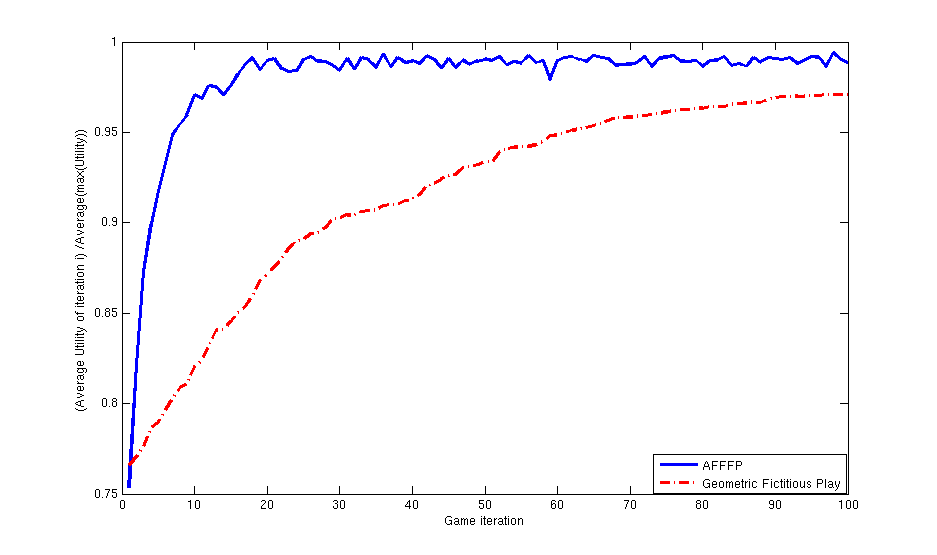}%
\caption{Utility of AFFFP (dotted line) and
geometric fictitious play (dashed line) for the vehicle target assignment game.}%
\label{fig:vta}%
\end{center}
\end{figure}

Figure \ref{fig:vta} depicts the average results for 100 instances of the game 
for the two algorithms, AFFFP and geometric fictitious play. For each instance, both algorithms run for 100 negotiation steps. To be able to average across the 100 instances we normalise the scores of an instance by the highest observed score for that instance (since some instances will have greater available score than others). As in the strategic form game, we set the randomisation parameter $\xi$ in the smooth best response function equal to 1 for both algorithms.

In Figure \ref{fig:vta} we observe that AFFFP result in a better solution on average than geometric fictitious play. Furthermore geometric fictitious play needs more iterations to reach the area where its reward is maximised than AFFFP.

%%%%%%%%%%%%disaster

\subsection{Disaster management scenario}
Finally we test our algorithm in a disaster
management scenario as described in \cite{stelios}. Consider the
case where a natural disaster has happened (an earthquake for
example) and because of this $N_{I}$ simultaneous incidents
occurred in different areas of a town. In each incident $j$, a
different number of people $N_{p}(j)$ are injured. The town has a
specific number of ambulances $N_{amb}$ available that are able to
collect the injured people. An ambulance $i$ can be at the area of
incident $j$ in time $T_{ij}$ and has capacity $c_{i}$. We
will assume that the total capacity of the ambulances is larger
than the number of injured people. Our aim is to allocate the
ambulances to the incidents in such a way that the average time
$\frac{1}{N_{amb}}\sum_{i:s^{i}=j}{T_{ij}}$ that the ambulances
need to reach the incidents is minimised while all the people
that are engaged in the incident will be saved. Then we can
formulate this scenario as follows. Each of the $N_{amb}$ players
should choose one of the $N_{I}$ incidents as actions. The
utility to the system of an allocation is:
\begin{equation}
u_{g}(s)=- \frac{1}{N_{amb}} \sum_{j=1}^{N_{I}}\sum_{i:
s^{i}=j}T_{ij}-\sum_{j=1}^{N_{I}}max\big(0,N_{p}(j)-\sum_{i:
s^{i}=j}c_{i}\big) \label{eq:disut}
\end{equation}
where $i=1,\ldots,N_{amb}$ and $s^{i}$ is the action of player
$i$. The emergency units have to ``negotiate'' with each other and
choose the incident which they will be allocated to, using a variant of fictitious play.

The first component of the utility function expresses the first
aim, to allocate the ambulance to an incident as fast as possible.
Thus the agents have to choose an incident with small $T_{ij}$.
The second objective, which is to save all the injured people that
are engaged in an incident, is expressed as the second component
of the utility function. It is a penalty factor that adds to the
average time the number of the people that were not able to be
saved. Like the vehicle target assignment game each player can
choose only one incident to help, but in each incident more than
one player can provide his help.

We follow \cite{stelios} and consider simulations with $3$ and $5$ incidents,
and $10$, $15$ and $20$ available ambulances. We run $200$ trials
for each of the combinations of ambulances and incidents, for each algorithm.
Since this scenario is \mbox{NP-complete} \cite{stelios} our aim is not to find
the optimal solution, but to reach a sufficient or a near
optimal solution. Furthermore the algorithm we present here is ``any-time'',
since the system utility generally increases as the time goes on, and therefore
interruption before termination results in good, if not optimal actions. In each
of the 200 trials the time $T_{ij}$ that an ambulance needs to
reach an incident is a random number uniformly distributed
between zero and one. The capacity of each ambulance is an
integer uniformly distributed between one and four. Finally the total
number of injured people that are involved in each incident is a
uniformly distributed integer between
$\frac{c_{t}}{2*N_{I}}$ and $\frac{c_{t}}{N_{I}}$, where $c_{t}$ is the
total capacity of the emergency units
$c_{t}=\sum_{i=1}^{Namb}c_{i}$.

In each trial we allow $200$ negotiation steps. In this scenario because of the utility function structure, a big randomisation parameter $\xi$ in the smooth best response function can easily lead to unnecessary randomisation. For that reason we set $\xi$ to $0.01$ for both algorithms which
results in a decision rule which approximates best response.
The learning rate and the initial value of $\lambda$ in adaptive forgetting
factor is set to $10^{-4}$ and $0.8$ respectively.

We use the same performance measures as \cite{stelios} to test the performance
of our algorithms. We compared the solution of our algorithm against
the centralised solution which can be obtained using binary
integer programming. In particular we  compared the solution of
our algorithm against the one we obtain by using Matlab's
\textit{bintprog} algorithm,which uses a branch and bound algorithm that is based on linear programming
relaxation \cite{bing1,bing2,bing3}. To compare the result of these two
algorithms we use the ratio $\frac{f_{fp}}{f_{opt}}$, where
$f_{fp}$ is the utility that the agents could gain
if they used the variations of fictitious play we propose and $f_{opt}$ is
the utility that the agents should gain if they were using
the solution of \textit{bintprog}. Thus values
of the ratio smaller than one mean that the proposed variations of fictitious
play perform better than \textit{bintprog}, and values of the ratio larger than
one mean that the proposed variations of fictitious play perform worst than
\textit{bintprog}. Furthermore we measured the
percentage of the instances in which all the casualties are
rescued, and the overall percentage of people that are rescued.

\begin{table}%
\begin{center}
\begin{tabular}{c| c| c| c| c}
\hline
&  &\scriptsize \%complete &\scriptsize \% saved &$f_{fp}/f_{opt}$ \\
\hline
& \scriptsize 10 ambulances& \scriptsize 82.0&\scriptsize 92.84&\scriptsize
1.2702 \\
\scriptsize 3 incidents& \scriptsize15 ambulances& \scriptsize 77.5&\scriptsize
89.88&\scriptsize 1.2624 \\
&\scriptsize 20 ambulances& \scriptsize 81.0&\scriptsize 90.59&\scriptsize
1.2058 \\
\hline
& \scriptsize10 ambulances&\scriptsize  91.0&\scriptsize 98.54&\scriptsize
1.6631 \\
\scriptsize 5 incidents& \scriptsize15 ambulances&\scriptsize  90.5&\scriptsize
98.28&\scriptsize 1.5088 \\
& \scriptsize20 ambulances& \scriptsize 83.0&\scriptsize 93.6&\scriptsize 1.4251
\\
\hline
\end{tabular}
\caption{Results of adaptive forgetting factor fictitious play after 200
negotiation steps for the three
performance measures.} \label{tab:dmres1}
\end{center}
\end{table}

\begin{table}%
\begin{center}
\begin{tabular}{c| c| c| c| c}
\hline
&  &\scriptsize \%complete &\scriptsize \% saved &$f_{fp}/f_{opt}$ \\
\hline
& \scriptsize 10 ambulances& \scriptsize 95,5&\scriptsize 99,74&\scriptsize
1.2970 \\
\scriptsize 3 incidents& \scriptsize 15 ambulances& \scriptsize 74,5
&\scriptsize 88.24&\scriptsize 1.2965 \\
&\scriptsize 20 ambulances& \scriptsize 60.55 &\scriptsize 87.9 &\scriptsize
1.2779 \\
\hline
& \scriptsize10 ambulances&\scriptsize  94.5 &\scriptsize 99.68&\scriptsize
1.8587 \\
\scriptsize 5 incidents& \scriptsize15 ambulances&\scriptsize  79.0&\scriptsize
88.28&\scriptsize 1.7443 \\
& \scriptsize20 ambulances& \scriptsize 48.50&\scriptsize 86.54&\scriptsize
1.8545 \\
\hline
\end{tabular}
\caption{Results of geometric fictitious play after 200 negotiation steps
for the three
performance measures.} \label{tab:dmres2}
\end{center}
\end{table}

Tables \ref{tab:dmres1} and \ref{tab:dmres2} present the results we have
obtained in the last step of negotiations between the ambulances
for the disaster management scenario when they use adaptive forgetting factor 
and geometric fictitious play respectively, to coordinate. 

The total percentage of the people that were saved and the ratio of
$f_{fp}/f_{opt}$ were similar within the groups of  3 and 5 incidents when the
adaptive forgetting factor fictitious play algorithm were used. Regarding the
percentage of the trials in which all people were saved, we can
observe that as we increase the complexity of the scenario, hence the number of
ambulances, the performance of adaptive forgetting factor fictitious play is
decreasing.

When we compare the results of the two algorithms we can observe that in both
cases of 3 and 5 incidents respectively adaptive forgetting factor fictitious
play perform better than geometric fictitious play when the scenarios included
more ambulances, and therefore were more complicated. Especially in the case of
the 20 ambulances the difference when we consider the number of the cases where
all the casualties were collected from the incidents, was greater than 20\%.

The differences we can observe from \textit{bintprog}'s centralised solution,
for both algorithms,
can be explained from the structure of the utility function (\ref{eq:disut}). The
first component of the utility is a number between zero and one
since it is the average of the times, $T_{ij}$, that the
ambulances need to reach the incidents. On the other hand the
penalty factor, even in the cases where only one person is not
collected from the incidents, is greater than the first component
of the utility. Thus a local search algorithm like the variations of fictitious
play we propose initially searches for an allocation that collects
all the injured people, so the penalty component of the utility
will be zero, and afterwards for the allocation that minimises
also the average time that the ambulances needed to reach the
incidents. It is therefore easy to become stuck in local optima.

We have also examined how the results are influenced by the number of
iterations we use in each of the 200 trials of the game. For
that reason we have compared the results that we would have obtained if in each
instance
of the simulations we had stopped the negotiations between the
emergency units after 50, 100, 150 and 200 iterations for both algorithms.
Tables
\ref{tab:dmrescoma}-\ref{tab:dmresrata} and
\ref{tab:dmrescom}-\ref{tab:dmresrat} depict the results for adaptive forgetting
factor fictitious play and geometric fictitious play respectively.

\begin{table}%
\begin{center}
\begin{tabular}{|c| c| c| c| c|c|}
\hline
& & \multicolumn{4}{c|}{Iterations} \\
\cline{3-6}
&  &50&100&150&200 \\
\hline
& \scriptsize 10 ambulances& 74.5&81.0 &82.0 &82.0 \\
\scriptsize 3 incidents& \scriptsize 15 ambulances& 73.0&75.5&77.5&77.5 \\
& \scriptsize 20 ambulances& 73.5&78.0&80.0&81.0 \\
\hline
&\scriptsize  10 ambulances& 80.0&87.5&90.0&91.0 \\
\scriptsize 5 incidents& \scriptsize 15 ambulances& 76.5&84.5&89.5&90.5 \\
& \scriptsize 20 ambulances& 62.0&77.0&80.0&83.0 \\
\hline
\end{tabular}
\caption{Percentage of solutions in which the capacity of the
ambulance in every incident was enough to cover all injured people
for different stopping times of the negotiations, 50, 100, 150 and
200 iterations of the adaptive forgetting factor fictitious play algorithm.}
\label{tab:dmrescoma}
\end{center}
\end{table}

\begin{table}%
\begin{center}
\begin{tabular}{|c| c| c| c| c|c|}
\hline
& & \multicolumn{4}{c|}{Iterations} \\
\cline{3-6}
&  &50&100&150&200 \\
\hline
&\scriptsize  10 ambulances& 91.44&92.35&92.95&92.84 \\
\scriptsize 3 incidents&\scriptsize  15 ambulances& 88.65&90.28&89.97&89.88 \\
&\scriptsize  20 ambulances& 89.09&89.33&90.44&90.59 \\
\hline
&\scriptsize  10 ambulances& 96.39&97.80&98.51&98.54 \\
\scriptsize 5 incidents&\scriptsize  15 ambulances& 94.84&97.22&98.03&98.29 \\
&\scriptsize  20 ambulances& 87.61&91.90&92.81&93.61 \\
\hline
\end{tabular}
\caption{Average percentage of injured people collected for
different stopping times of the negotiations, 50, 100, 150 and 200
iterations of the adaptive forgetting factor fictitious play algorithm.}
\label{tab:dmressava}
\end{center}
\end{table}

\begin{table}%
\begin{center}
\begin{tabular}{|c| c| c| c| c|c|}
\hline
& & \multicolumn{4}{c|}{Iterations} \\
\cline{3-6}
&  &50&100&150&200 \\
\hline

& \scriptsize 10 ambulances& \scriptsize 1.2791&\scriptsize 1.2603&\scriptsize
1.2669&\scriptsize 1.2702 \\
\scriptsize 3 incidents&\scriptsize  15 ambulances& \scriptsize
1.2701&\scriptsize 1.2452&\scriptsize 1.2319&\scriptsize 1.2624 \\
&\scriptsize  20 ambulances&\scriptsize  1.2142&\scriptsize 1.1971&\scriptsize
1.1943&\scriptsize 1.2058 \\
\hline

&\scriptsize  10 ambulances& \scriptsize 1.6989&\scriptsize 1.6827&\scriptsize
1.6772&\scriptsize 1.6631 \\
\scriptsize 5 incidents&\scriptsize  15 ambulances&\scriptsize 
1.5569&\scriptsize 1.5352&\scriptsize 1.5304&\scriptsize 1.5088 \\
&\scriptsize  20 ambulances&\scriptsize 1.5298&\scriptsize 1.4413&\scriptsize
1.4306&\scriptsize 1.4251 \\
\hline
\end{tabular}
\caption{Average percentage of the ratio $f_{fp}/f_{opt}$ for
different stopping times of the negotiations, 50, 100, 150 and 200
iterations of the adaptive forgetting factor fictitious play algorithm.}
\label{tab:dmresrata}
\end{center}
\end{table}

\begin{table}%
\begin{center}
\begin{tabular}{|c| c| c| c| c|c|}
\hline
& & \multicolumn{4}{c|}{Iterations} \\
\cline{3-6}
&  &50&100&150&200 \\
\hline
& \scriptsize 10 ambulances& 94.0&94.0 &94.7 &95.5 \\
\scriptsize 3 incidents& \scriptsize 15 ambulances& 71.0&73.0&73.3&74.5 \\
& \scriptsize 20 ambulances& 60.5&61.3&62.0&63.0 \\
\hline
&\scriptsize  10 ambulances& 86.0&92.0&93.3&94.5 \\
\scriptsize 5 incidents& \scriptsize 15 ambulances& 79.0&78.0&79.0&82.0\\
& \scriptsize 20 ambulances& 42.0&49.0&47.3&48.5 \\
\hline
\end{tabular}
\caption{Percentage of solutions in which the capacity of the
ambulance in every incident was enough to cover all injured people
for different stopping times of the negotiations, 50, 100, 150 and
200 iterations of the geometric fictitious play algorithm.}
\label{tab:dmrescom}
\end{center}
\end{table}

\begin{table}%
\begin{center}
\begin{tabular}{|c| c| c| c| c|c|}
\hline
& & \multicolumn{4}{c|}{Iterations} \\
\cline{3-6}
&  &50&100&150&200 \\
\hline
&\scriptsize  10 ambulances& 99.62&99.65&99.69&99.74 \\
\scriptsize 3 incidents&\scriptsize  15 ambulances& 88.20&88.21&88.21&88.24 \\
&\scriptsize  20 ambulances& 86.65&87.41&88.23&89.90 \\
\hline
&\scriptsize  10 ambulances& 97.20&99.54&99.61&99.68 \\
\scriptsize 5 incidents&\scriptsize  15 ambulances& 94.84&97.22&98.03&98.29 \\
&\scriptsize  20 ambulances& 85.33&86.38&86.391&86.54 \\
\hline
\end{tabular}
\caption{Average percentage of injured people collected for
different stopping times of the negotiations, 50, 100, 150 and 200
iterations of the geometric fictitious play algorithm.} \label{tab:dmressav}
\end{center}
\end{table}

\begin{table}%
\begin{center}
\begin{tabular}{|c| c| c| c| c|c|}
\hline
& & \multicolumn{4}{c|}{Iterations} \\
\cline{3-6}
&  &50&100&150&200 \\
\hline

& \scriptsize 10 ambulances& \scriptsize 1.2957&\scriptsize 1.2928&\scriptsize
1.3039&\scriptsize 1.2970 \\
\scriptsize 3 incidents&\scriptsize  15 ambulances& \scriptsize
1.2965&\scriptsize 1.3039&\scriptsize 1.2801&\scriptsize 1.2740 \\
&\scriptsize  20 ambulances&\scriptsize  1.2779&\scriptsize 1.2744&\scriptsize
1.2723&\scriptsize 1.2722 \\
\hline

&\scriptsize  10 ambulances& \scriptsize 1.8587&\scriptsize 1.8540&\scriptsize
1.8550&\scriptsize 1.8519 \\
\scriptsize 5 incidents&\scriptsize  15 ambulances&\scriptsize 
1.7443&\scriptsize 1.7205&\scriptsize 1.7085&\scriptsize 1.7074 \\
&\scriptsize  20 ambulances&\scriptsize 1.8545&\scriptsize 1.7845&\scriptsize
1.7744&\scriptsize 1.7727 \\
\hline
\end{tabular}
\caption{Average percentage of the ratio $f_{fp}/f_{opt}$ for
different stopping times of the negotiations, 50, 100, 150 and 200
iterations of the geometric fictitious play algorithm.} \label{tab:dmresrat}
\end{center}
\end{table}

We can see from tables \ref{tab:dmrescoma}-\ref{tab:dmresrata} that  the
performance of adaptive forgetting factor fictitious play, in all the measures
that we used, is similar for 100, 150 and 200 negotiation steps. In particular
when we consider the percentage  of the instances that the ambulances collected
all the injured people  and the ratio $f_{fp}/f_{opt}$ the difference in
the results after 100 and 200 negotiation steps is between
1\% and 4.5\%. The differences become even smaller for the percentage of the
people that were saved which was less than 2\%. 

Geometric fictitious play was trapped in an area of a local minimum after few
iterations since the results are similar after 50, 100, 150 and 200 iterations.
This is reflected in the results where geometric fictitious play performed worse
than  adaptive forgetting factor fictitious play especially in the complicated
cases where the negotiations where between 20 ambulances.

Adaptive forgetting factor fictitious play performed also better than the Random Neural Network (RNN)
presented in \cite{stelios}, when we consider
the percentage of the cases in which all the injured people are collected
and the overall percentage of people that are rescued. The
percentage of instances where the proposed allocations by the
RNN could collect all the casualties were from $25$ to $69$
percent. The corresponding results of adaptive forgetting factor fictitious play
are from  $77.5$ to $94.5$. The overall
percentage of people that are rescued by the RNN algorithm are
similar to the ones of adaptive forgetting factor fictitious play, between $85$
and $98.5$ percent. The ratio $\frac{f_{fp}}{f_{opt}}$ reported by \cite{stelios} is better than that shown here. However in \cite{stelios} only the examples in which all the casualties were collected were included
to evaluate the ratio. Cases with high penalties, since the
uncollected casualties introduce higher penalties than the inefficient
allocation, were excluded from the ratio evaluation. Thus artificially improve their metric, especially when one considers that in many instances less than 40\% of their solutions were included.

%%%%% Conclusions
\section{Conclusions}
Fictitious play is a classic learning
algorithm in games, but it is formed on an (incorrect)
stationarity assumption. Therefore we have introduced
a variation of fictitious play, adaptive forgetting factor fictitious play,
which address this problem by giving higher weights to the recently observed
actions using a heuristic rule from the streaming data literature.

We examined the impact of adaptive forgetting factor fictitious play parameters
$\lambda_{0}$ and $\gamma$ on the results of the algorithm. 
We showed that these two parameter should be chosen carefully since there are
combinations of $\lambda_{0}$ and $\gamma$ that induce very poor results. An
example of such combination is  when high values of the learning rate $\gamma$
are combined with low values of $\lambda_{0}$. This is because values of $\lambda_{0}<0.6$ assign small weights to the previously observed actions and this results in volatile estimations that are influenced by opponents' randomisation. High values of the learning rate $\gamma$, mean that $\lambda_0$ is driven still  lower, exacerbating the problem further. From the simulation results we have seen that a satisfactory combination of parameters $\lambda_0$ and $\gamma$ is $0.8 \leq \lambda_0 \geq 0.9$ and $\gamma=10^{-4}$.

Adaptive forgetting factor performed better than the competitor algorithms in the climbing hill game. Moreover it converged
to the a better solution than geometric fictitious play  in the vehicle target assignment game. In the disaster
management scenario the performance of
the proposed variation of fictitious play compared favorably with that of geometric
fictitious play and a pre-planning algorithm that uses neural networks \cite{stelios}. 

Our empirical observations indicate that adaptive forgetting factor fictitious play converges to a
solution that is at least as good as that given by the competitor algorithms. Hence by
slightly increasing the computational intensity of fictitious play less communication
is required between agents to quickly coordinate on a desirable solution.

%\begin{acknowledgements}
%Michalis Smyrnakis is supported by The Engineering and Physical 
%Sciences Research Council EPSRC (grant number EP/I005765/1).
%\end{acknowledgements}

\bibliographystyle{splncs}      % mathematics and physical sciences
\bibliography{AFFFP}   % name your BibTeX data base

\end{document}